\documentclass[11pt, twocolumn, letterpaper]{article}

\usepackage[table]{xcolor}
\usepackage{wacv}

\usepackage{times}
\usepackage{epsfig}
\usepackage{graphicx}
\usepackage{amsmath}
\usepackage{lmodern}

\usepackage[stable]{footmisc}
\usepackage{amssymb}
\usepackage{caption}
\usepackage{subcaption}
\usepackage{wrapfig}
\usepackage{siunitx}
\usepackage{dblfloatfix} 
\usepackage{tabularx}
\usepackage{lipsum}
\usepackage{floatrow}
\usepackage[normalem]{ulem}

\useunder{\uline}{\ul}{}
\usepackage{titling}

\wacvfinalcopy % *** Uncomment this line for the final submission

\ifwacvfinal\fi
\author{Darius Lam\thanks{D. Lam is at Harvard College, in support of Defense Innovation Unit Experimental (DIUx)}
\and
Richard Kuzma\thanks{R. Kuzma and B. McCord are at DIUx}
\and 
Kevin McGee\thanks{K. McGee is at DigitalGlobe}
\and 
Samuel Dooley\thanks{S. Dooley, M. Laielli, and M. Klaric are at the National Geospatial-Intelligence Agency (NGA)}
\and Michael Laielli\footnotemark[4]
\and Matthew Klaric\footnotemark[4]
\and Yaroslav Bulatov\thanks{Y. Bulatov is in support of DIUx}
\and Brendan McCord\footnotemark[2]
}

\title{xView: Objects in Context in Overhead Imagery}
\thanksmarkseries{arabic}
\date{}

\begin{document}

%%%%%%%%% TITLE

% Authors at the same institution
%\author{First Author \hspace{2cm} Second Author \\
% Authors at different institutions

\maketitle

\ifwacvfinal\thispagestyle{plain}\fi

%%%%%%%%% ABSTRACT
\begin{abstract}

We introduce a new large-scale dataset for the advancement of object detection techniques and overhead object detection research.  This satellite imagery dataset enables research progress pertaining to four key computer vision frontiers.% reducing the minimum resolution required to achieve certain results; improving learning efficiency; pushing the limit on how many object classes can be discovered; improving the ability to detect fine-grained classes. % 
We utilize a novel process for geospatial category detection and bounding box annotation with three stages of quality control.  Our data is collected from WorldView-3 satellites at 0.3m ground sample distance, providing higher resolution imagery than most public satellite imagery datasets.  We compare xView to other object detection datasets in both natural and overhead imagery domains and then provide a baseline analysis using the Single Shot MultiBox Detector.  xView is one of the largest and most diverse publicly available object-detection datasets to date, with over 1 million objects across 60 classes in over 1,400 km$^2$ of imagery.

\end{abstract}

\begin{figure}
\centering
\includegraphics[width=\linewidth]{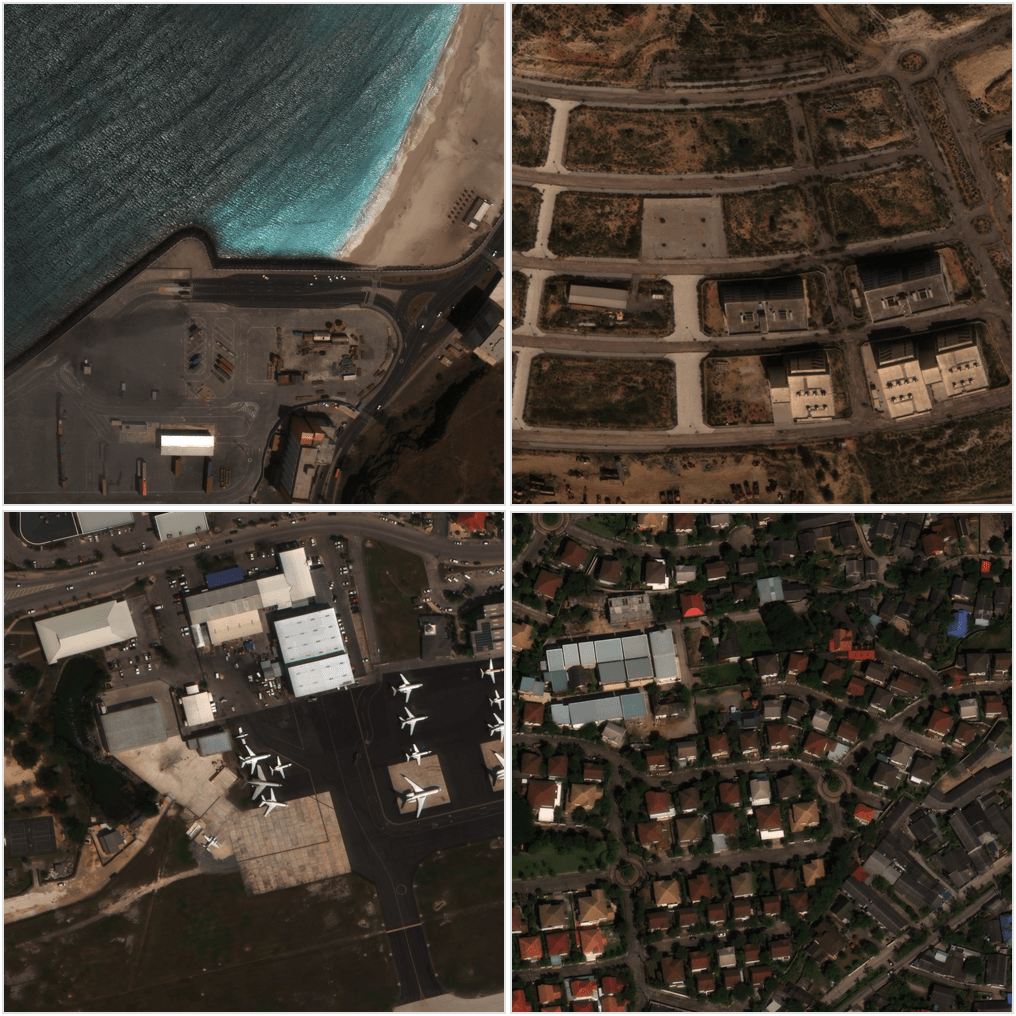}
\caption{Four of the many views of xView.  Imagery comes from different geographical locations with different levels of human use.  Imagery in this figure is from DigitalGlobe.}
\end{figure}

%%%%%%%%% BODY TEXT
\section{Introduction} 

The abundance of overhead image data from satellites and the growing diversity and significance of real-world applications enabled by that imagery provide impetus for creating more sophisticated and robust models and algorithms for object detection. We hope xView will become a central resource for a broad range of research in computer vision and overhead object detection.

The vast majority of satellite information in the public domain is unlabeled. Azayev notes a lack of labeled imagery for developing deep learning methods \cite{Azayev}.  Hamid et al. also note that the curation of high quality labeled data is central to developing remote sensor applications \cite{Hamid}.  Ishii et al., Chen et al., and Albert et al. all develop methods for detection or segmentation of buildings, all using different (and sometimes custom collected and labeled) datasets \cite{Ishii,Chen,Albert}.  The utilization of different datasets makes it difficult to compare results between different authors.  We developed xView as a general purpose object detection dataset of satellite imagery so as to be familiar to the computer vision community and remote sensing community alike.

Several object detection datasets exist in the natural imagery space, but there are few for overhead satellite imagery. The public overhead datasets in existence typically suffer from low class count, poor geographic diversity, few training instances, or too narrow class scope. xView remedies these gaps through a significant labeling effort involving the collection of imagery from a variety of locations and the use of an ontology of parent- and child-level classes.

We created xView with four computer vision frontiers in mind:

\textbf{Improve Minimum Resolution and Multi-Scale Recognition:} Computer vision algorithms often struggle with low-resolution objects \cite{Fang2015,lrcnn}. For example, the YOLO architecture limits the number of predictable bounding boxes within a spatial region, making it difficult to detect small and clustered objects \cite{yolo}.  Detecting objects on multiple scales is an ongoing topic of research \cite{hongyang,zhaowei}. Objects in xView vary in size from 3 meters (10 pixels at 0.3 meters ground-sample distance [GSD]) to greater than 3,000 meters (10,000 pixels at 0.3 meters GSD).  The varying ground sample distance of different satellites means that xView has significantly higher resolution than many public satellite imagery datasets.

\textbf{Improve Learning Efficiency:}    In the real world, objects are often not evenly distributed within images.  There may be many thousands more cars in any given city than there are hospitals.  Imbalanced classification and localization on uneven datasets is important for real-world applications.  xView captures this property by including object classes with few instances as well as classes with many instances (see Figure 4). 

\textbf{Push the Limit of Discoverable Object Classes:}   xView includes 60 classes.  For reference, COCO includes 91 classes and SpaceNet includes 2 classes.  There is significant class diversity in xView, which contains both land-use and pedestrian classes with easily discretized objects such as cars and buildings as well object classes involving groupings of multiple object types such as construction sites and vehicle lots.  

\textbf{Improve Detection of Fine Grained Classes:} Fine grained object detection is necessary for practical applications. Detecting a 'sailboat' gives different information than detecting an 'oil tanker', despite them both being 'maritime vessels'. Fine grained object detection is a difficult task and an ongoing area of research \cite{Zhao2017, Turner2015, Karlinsky_2017}.  Over 80\% of classes in xView are fine grained, belonging to 7 different parent classes. For example, xView contains 8 distinct truck child classes, including 'pickup truck', 'utility truck', and 'cargo truck'.

To create xView we designed a substantive annotation and quality control process. With chipped satellite images delivered in RGB and 8-band multispectral format, annotators used QGIS (Q-Geographic Information System), an open source tool, to load up and mark image chips.  Using an in-house plugin, annotators are able to create axis-aligned bounding boxes for individual objects.  Our dataset includes images prepared in ways that are typical for satellite images including orthorectification, pan-sharpening, and atmospheric correction. 

In order to minimize biased image sampling, we define scene types that are relevant to many overhead applications and strive for a uniform distribution of images across those scene types as well as the ways in which those scenes may appear. Variability in scene type can come from a place's function, while the visual appearance of a scene may vary according to a multitude of factors.  xView pulls from a wide range of geographic locations (see Figure 11). Each location has its own distinct features, including physical differences (desert, forest, coastal, plains) and constructional differences (layout of houses, cities, roads). The variety of collection geometries possible with satellite imagery produces images with multiple perspectives on objects within a given class.

The xView dataset contains 60 object categories with 1 million labeled objects covering over 1,400 km$^2$ of the earth's surface.  The large chip sizes allow variability in pre-processing techniques; we discuss several options in section 4. xView has a similar number of instances and class counts as COCO and substantially greater number of classes than SpaceNet and Cars Overhead with Context \cite{coco,spacenet,cowc}.
%Show COWC labeled imagery, SpaceNet imagery, and xView side by side

\begin{figure*}
\centering
\includegraphics[width=\linewidth]{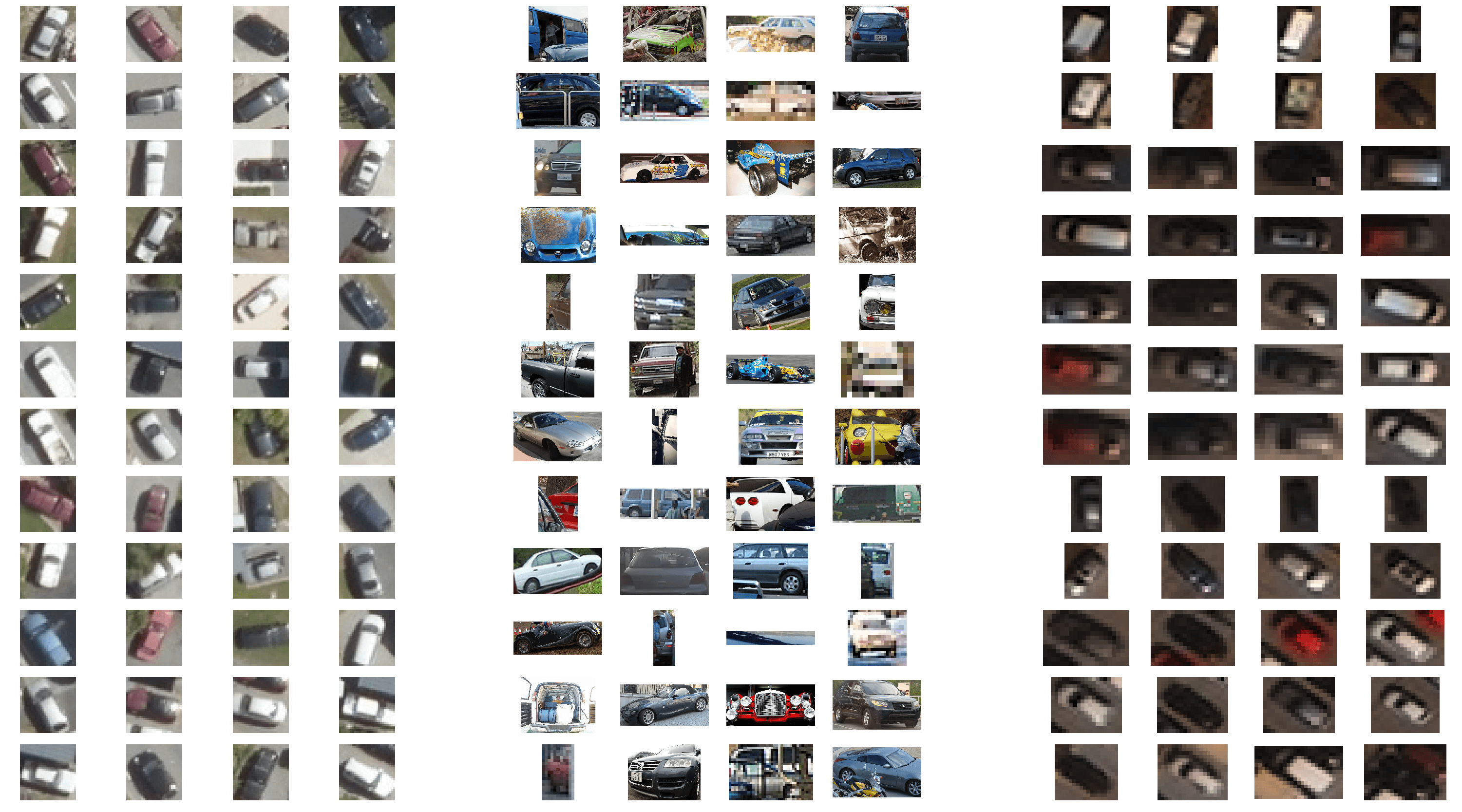}
\caption{COWC, PASCAL VOC, and xView cars, respectively \cite{cowc, pascalvoc}.  COWC provides object labels as single points corresponding to the center of each car \cite{cowc}.  To generate bounding boxes, we create 20x20 pixel boxes around each point. COWC and xView imagery in this figure are extracted from a single location.  xView imagery in this figure is from DigitalGlobe.}
\end{figure*}

\section{Related Work}

The complexity of our world, combined with differences in collection geometry from space-based imaging platforms, makes object recognition in satellite imagery a difficult task. xView contributes a large, multi-class, multi-location dataset in the object detection and satellite imagery space, built with the benchmark capabilities of PASCAL VOC, the quality control methodologies of COCO, and the contributions of other overhead datasets in mind.  This combination opens up opportunities for applied research in census mapping (e.g., correlating building count and inhabitance), economic reporting (e.g., predicting income level through vehicle density), disaster response (e.g., identifying damaged regions), and more.  

The task of object detection is properly identifying an object within an image and localizing it, either through bounding boxes or segmentation. One such dataset, PASCAL VOC, has been maintained since 2005 and has grown to 20 object classes in 11,530 images containing 27,450 bounding boxes and ~7,000 segmentations \cite{pascalvoc}.  In the past decade, the PASCAL VOC dataset has been widely used in the object detection space.  A number of object detection papers have used PASCAL VOC as a benchmark \cite{yolo, coco, fasterrcnn, rfcn}. PASCAL VOC, however, contained mostly ``iconic view'' scenes that are often non-representative of the real world, an issue remedied by COCO. COCO contains 91 object classes across around 328,000 images.  The COCO dataset has on average more categories per image at smaller sizes than PASCAL VOC \cite{coco}.  The ImageNet detection dataset is a large-scale dataset containing 200 classes and around .5 million labeled instances (ILSVRC 2014) \cite{imagenet}. Most recently, Google released OpenImages V2, a large-scale dataset containing 1.6 million images, around 4 million bounding boxes, and 600 object classes on natural imagery \cite{openimages}.

The Cars Overhead with Context (COWC) dataset by the Lawrence Livermore National Laboratory is an overhead image dataset with around 32,700 labeled cars \cite{cowc}.  COWC uses aerial image capture as opposed to satellite image capture, so their images are of high resolution but capture less area.  COWC includes images from six locations.  The SpaceNet dataset focuses on object segmentation.  SpaceNet has segmentation masks for around 5 million buildings in 5 locations \cite{spacenet}.  Recently, the SpaceNet dataset has expanded to include roads.  Both SpaceNet and COWC have images taken at similar times of day. Each of these datasets contains few classes in few geographic regions, limiting their usability for general object recognition in overhead imagery.  Recently, IARPA released their Functional Map of the World (FMoW) dataset with RGB and multispectral imagery for recognizing functional land use from temporal sequences of satellite images \cite{fmow}. FMoW contains around 1 million images in 63 categories from over 200 countries.  The FMoW dataset is designed for temporal reasoning in classification of land-use subregions. FMoW classes do not include vehicles (e.g., sailboat, fishing vessel, and small car) \cite{fmow}. xView includes vehicles, which makes it more representative of the real world and also better targets the multi-scale problem. 

Figure 2 highlights the differences between xView, PASCAL VOC, and COWC car classes.  COWC only contains cars, while xView also contains other vehicles like trucks and tractors. COWC images are also only captured aerially, so cars are more consistent in viewpoint and lighting.  PASCAL VOC contains other non-car vehicular classes, such as bus and train, but all images are all ground-level natural imagery.  VOC cars also occupy a larger proportion of area than COWC and xView cars.  xView cars have varied sensor elevation levels and are from more locations than COWC.  xView cars provide a better indication of environments that would be experienced in real life: not all cars in satellite images will be imaged from a 90-degree elevation angle and in perfect daylight.  
%need to accentuate the differences between xview & other
\section{Dataset Details}

 xView is one of the largest and most diverse publicly available overhead imagery datasets for object detection. Annotators used QGIS (Q-Geographic Information System), an open-source satellite imagery manipulation tool, along with a developed in-house plugin to create axis-aligned bounding boxes for individual objects.  Our dataset includes images prepared in ways that are typical for satellite images including orthorectification, pan-sharpening, and atmospheric correction.  

\subsection{Image Collection}

We created xView by first selecting a wide group of object categories to be considered.  Of this group, we down-selected to 60 classes, which were organized in a parent-child format where parents were more general categorizations and children represented specific instances of these general categories.  For example, the `engineering vehicle' parent class contained the `excavator' child class.  xView contains seven parent classes: `fixed wing aircraft', `passenger vehicle', `truck', `railway vehicle', `engineering vehicle', `maritime vessel', and `building'. Not all classes are contained under a parent superset (e.g., 'helipad'). 

In order to minimize biased image sampling, we select imagery from an appropriate distribution of areas of interest (AOIs). AOIs in xView include mines, ports, airfields, and coastal, inland, urban, and rural regions.  They also exist over multiple continents. AOI selection can be broken down into two smaller steps: identifying lat-long coordinates for investigation and drawing polygons in larger areas surrounding those points.  First, lat-long coordinates of  mines, ports, airfields, and other locations with relevant classes are collected from open source databases. Next, we determine if satellite imagery covers the lat-long coordinate and surrounding area.   A manual search is performed to confirm the existence of objects in the relevant class and if present, a polygon is drawn, creating the AOI to be extracted from the larger satellite imagery available in the region. 

The next step is to collect 1 km$^2$ image chips based on the specified AOIs. This is done in two parts: selecting the image strips and applying a 1 km$^2$ grid of the polygons that intersect previously collected imagery. Some may have multiple image strip options and may be chosen in order to vary weather conditions (snow, clouds, etc.) or to capture ephemeral construction sites. Once an image strip is chosen, a 1 km$^2$ grid is applied. All grid cells that are intersected by a specified polygon are chipped out. The 1 km$^2$ grid is a Universal Transverse Mercator (UTM) zone derived grid to ensure 1 km$^2$ of area per grid cell, and to establish a repeatable process. The grid also serves as a production tool, to keep crowd workers focused on a small area for feature extraction, and to mark areas as complete by object type to track progress.

\subsection{Image Annotation}

The design of a high quality annotation pipeline was critical for xView. This pipeline included a crowdsourcing approach that leveraged labelers with experience across the breadth of classes in the dataset. We achieved consistency by having all annotation performed at a single facility, following detailed guidelines, with output subject to multiple quality control checks. Workers extensively annotated image chips with bounding boxes using an open source tool.

%Show QGIS screenshot

\begin{figure}
\centering
\includegraphics[width=\linewidth]{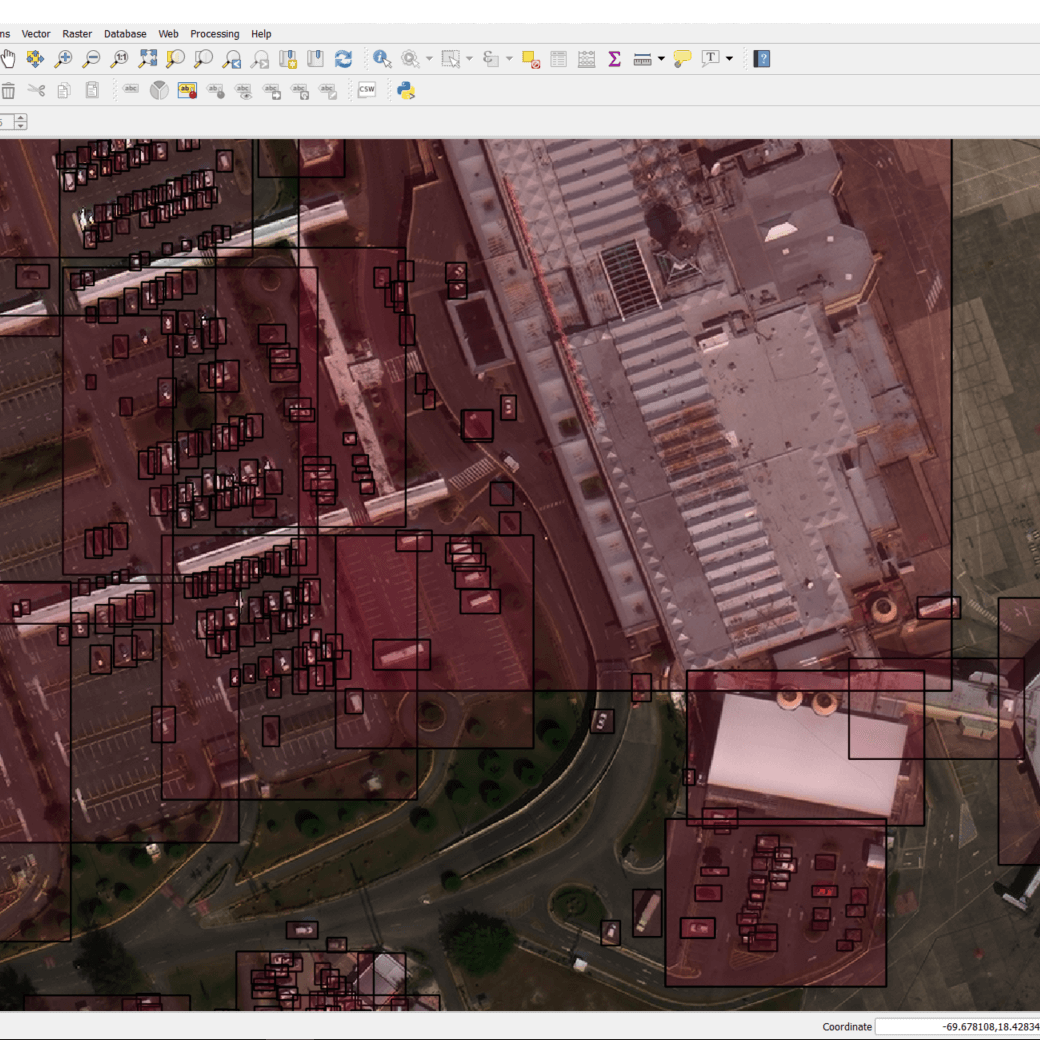}
\caption{The QGIS annotation software.  Drawn annotations are shown in red. Imagery in this figure is from DigitalGlobe.}
\end{figure}

%Add Instance Count Chart
%\begin{figure*}
%\centering
%\includegraphics[width=\linewidth]{"Instances Per  Category_small".png}
%\caption{xView instance count distribution by class}
%\label{fig:short}
%\end{figure*}

%Add Average Pixel Area Chart
%\begin{figure*}
%\includegraphics[width=\linewidth]{"box-whisker_small".png}
% \caption{xView pixel area distribution by class}
%\label{fig:short2}
%\end{figure*}

\begin{figure*}
\begin{subfigure}{\linewidth}
  \centering
  \includegraphics[width=\linewidth]{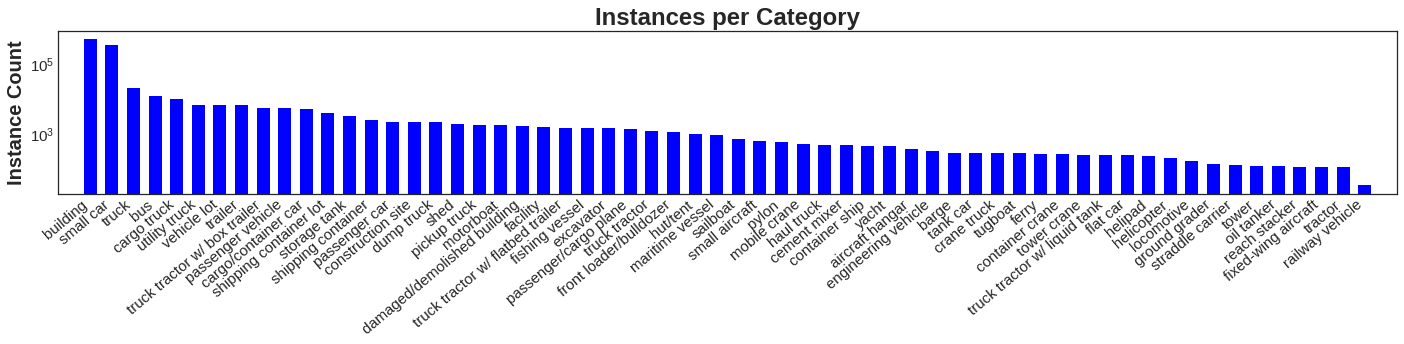}
\end{subfigure}
\vspace{10pt}
\begin{subfigure}{\linewidth}
  \centering
  \includegraphics[width=\linewidth]{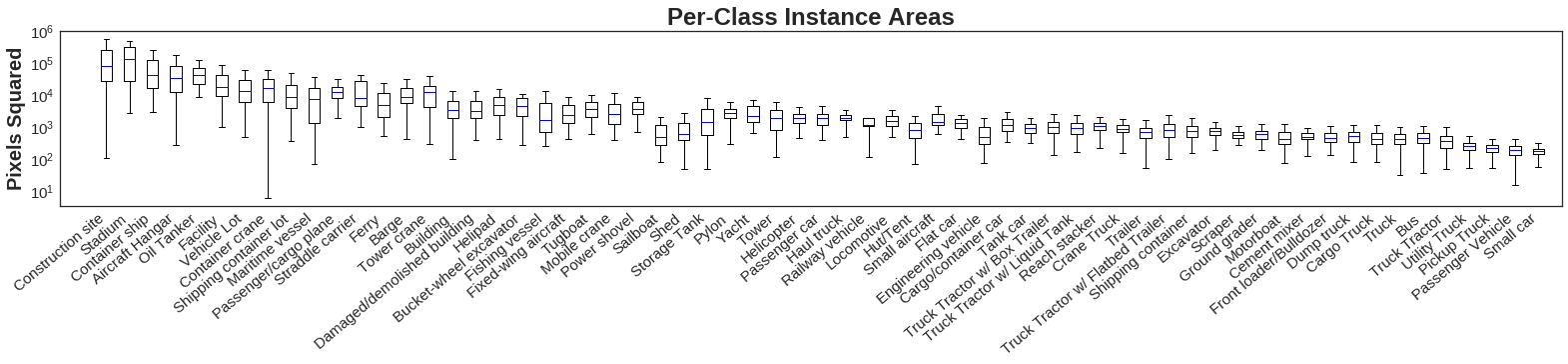}
\end{subfigure}
\caption{Top: xView instance count distribution by class.  Bottom: xView pixel area distribution by class.}
\end{figure*}

The annotators were trained through multiple sessions, both online and in person, in which they saw examples of the total 60 classes using both cross view and overhead imagery with both correct and incorrect labels.  Annotators were instructed to draw axis-aligned bounding boxes that tightly contained the object instance.  The private crowd had direct messaging access to the dataset team. Questions (e.g., ``What differentiates a facility from a building?'') could be relayed and solved in a timely manner.
                   
Each labeler is responsible for labeling all instances of one general category. If the labeler can confidently determine an instance to be of a fine grained child category, then that labeler will classify the instance at the child level. However, if the labeler is unable to make a confident determination, then they will fall back to the more general parent category.  Annotators are instructed to annotate all objects within the chip, unless the object is approximately $\leq20\%$ visible. Bounding box annotation is completed one 1 km$^2$ chip at a time, to keep work focused on small areas, track completion by object class, and provide discrete areas to the quality control reviewers to provide immediate feedback to the labeling team.  For clusters of objects, annotators were instructed to label each object individually.  For example, clusters of tightly packed buildings were present in some images, and each building was to be individually labeled to avoid labeling object groups. In practice, this was difficult to achieve, and groups of objects could be labeled as singular due to human error.  In large-scale overhead imagery, this error is representative of data encountered in real-world situations.

\subsection{Quality Control and Gold Standards}

Annotation quality control was conducted in three stages: Worker, supervisory, and expert. Worker quality control involved annotators performing the role of quality control reviewer on a rotational basis so that they could check the work of others, identify errors, and improve their own annotation. Reviews focused on category identification, bounding box size, bounding box orientation, and duplication. 

The second stage, supervisory quality control, involved checks for duplicate features, invalid labels, invalid geometries (polygons rather than axis-aligned bounding boxes), non-exhaustively labeled image chips, features that fell outside of image chips, and empty tiles.  The supervisory step produced feedback in the form of worker training sessions in addition to maintaining quality.

The third stage, expert quality control, involved creating a gold standard dataset and analyzing worker quality by applying a precision and recall threshold measurement between batches of worker-produced annotations and that gold standard dataset. Gold data was created by sampling and labeling six 1 km$^2$ chips from each batch by expert workers who were co-authors of the paper as well as professional imagery analysts. The batches represented 10/40/70/100\% of the total dataset. In order to pass expert quality control, the batch was required to have a precision of 0.75 and recall of 0.95 at 0.5 intersection over union (IoU) when compared to the gold standard. Batches that failed expert quality control were remediated and resubmitted. We utilized an additional crowd-sourcing platform called Tomnod for remediation of difficult chips. 

\subsection{Dataset Statistics}

%Add Number of Categories vs Number of Instances (see MSCOCO)

\begin{figure}
\centering
\includegraphics[width=\linewidth]{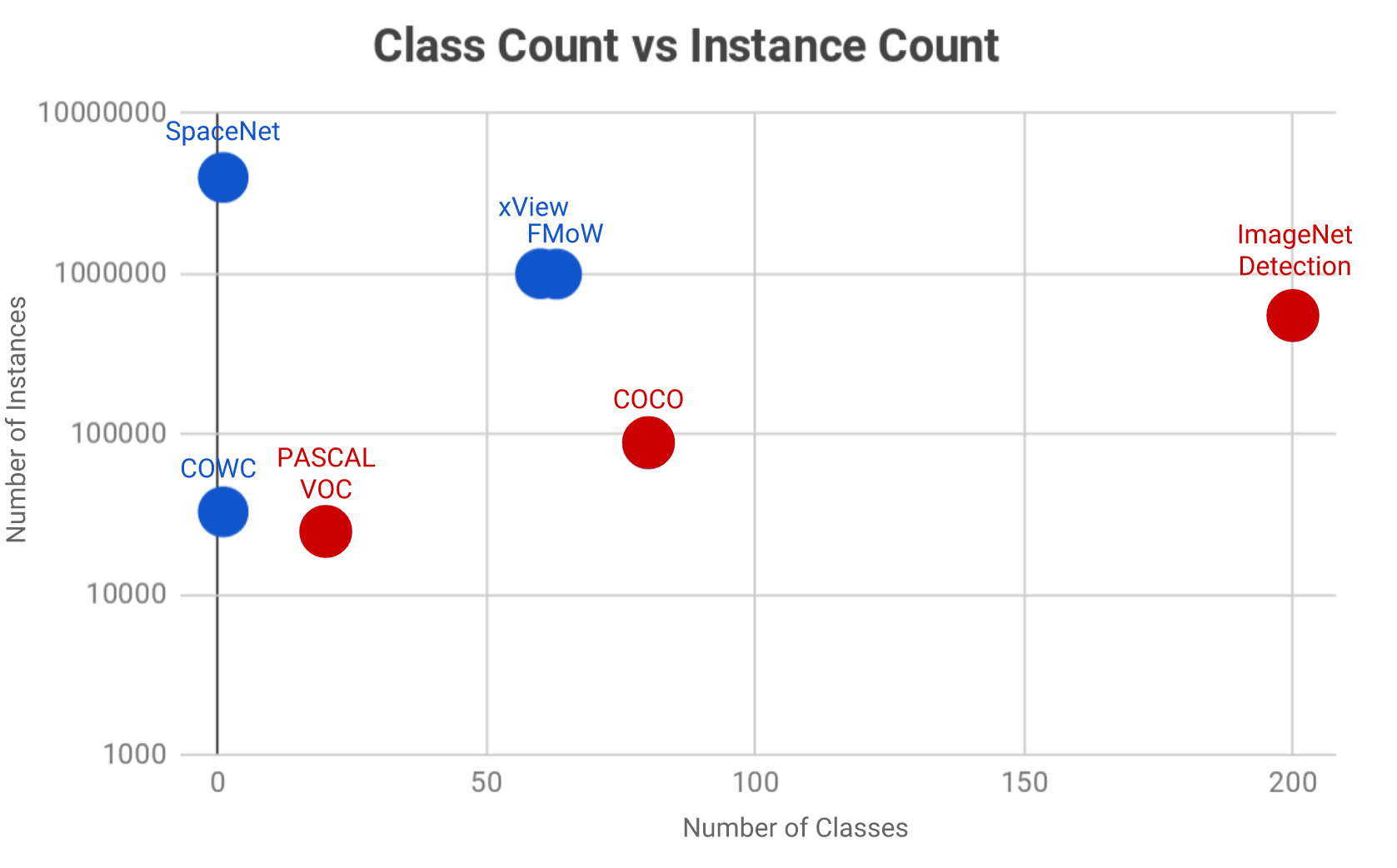}
\caption{Total Instance Count versus Number of Classes for major object detection datasets.  Blue indicates overhead imagery datasets; red indicates natural imagery datasets.}
\end{figure}

The xView dataset covers over 1,400 km$^2$ of the earth's surface, with 60 classes and approximately 1 million labeled objects. The most common classes are `building' and `small car' due to their prevalence in densely populated areas. Figure 4 shows the number of instances per class as well as the average pixel area per class. Analogous with their real-world prevalence, buildings and small cars have the highest instance counts. By average pixel area, the largest objects are typically land-use locations (e.g., 'construction sites', 'marinas', 'facilities', 'vehicle lots').  

%Add Train-Val competition split @SamDooley @RichardKuzma
We made three splits to the public release of xView: train, test, and val.  The total percentages of objects in each split are 59.2\%, 20.0\%, and 20.8\% for train, test, and val, respectively. Each split has at least 5 instances of each category, which is especially important for categories with a small number of instances.  We created splits on a per-image level, which made it difficult to proportionally split by category since xView images are large and object categories are spatially correlated.  The average of the per-category percentages in each split is 60.0\%, 21.0\%, and 19.4\% for train, test, and val, respectively.

Additionally, we compare xView to other object detection datasets.  Figure 5 shows total number of classes versus total number of instances in several object detection datasets. The closest dataset to xView to date is the FMoW land use dataset.  SpaceNet has a greater number of instances but only for buildings and roads. xView incorporates images from various countries. Each location has distinct visual features that affects the appearances of objects within that location (see Figure 6). This variety will drive a need for contextual understanding and adaptability to multiple view-types of objects.

\begin{figure}
\centering
\includegraphics[width=150px]{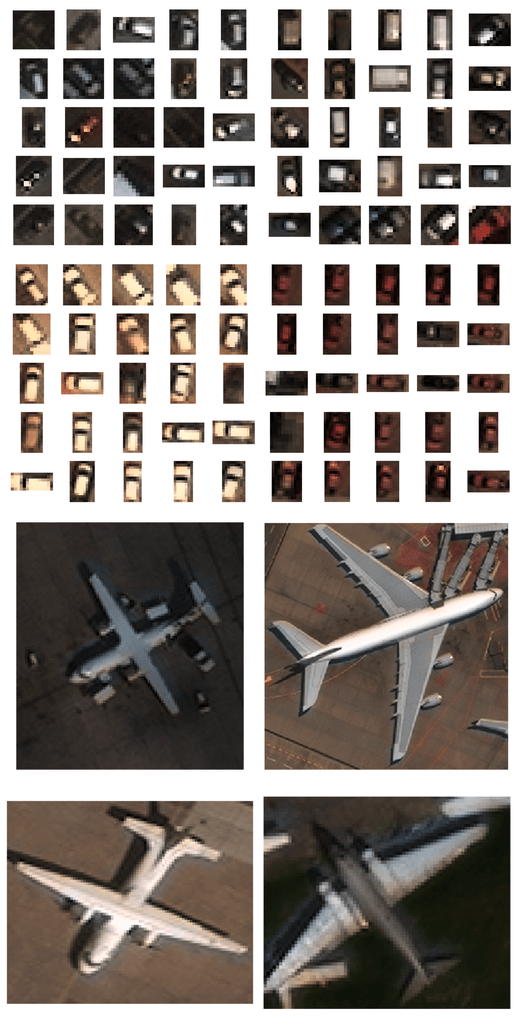}
\caption{Top: Extracted cars from four separate geographic locations. Bottom: Extracted planes four geographic locations. Imagery in this figure is from DigitalGlobe.}
\end{figure}

\section{Algorithmic Analysis}

We conducted object detection experiments using the Single Shot Multibox Detector meta-architecture (SSD).  SSD extracts prediction features at multiple layers for better multi-scale detection \cite{ssd}.  We evaluated three permutations of the dataset in order to assess xView difficulty and establish a baseline for future research. Because each 1 \si{km\square} image is around $3,000^2$ pixels, we first pre-processed the data by chipping it into $300^2$ pixel non-overlapping images. The large input images allow for great range of chipping options.  For our experiments, bounding boxes partially overlapped with a chip were cropped at the chip edge.  This infrequently resulted in cropping sections of large objects like facilities and stadiums.  We then created a multi-resolution dataset by chipping at different sizes (300$^2$, 400$^2$, 500$^2$), a pre-processing step suggested by Zhang et al \cite{zhang07}.  We also created a multi-resolution-augmented dataset via added image augmentation onto the multi-resolution dataset (shifting, rotation, noise, and blurring) (see Figure 7). We evaluated the SSD method over the three datasets: vanilla, multi-resolution (multires), and multi-resolution-augmented (aug).  We ran experiments on four M60 GPUs for seven days.  Each dataset was created by splitting the entire xView dataset into 70-30\% split for train and test, respectively.  All three datasets were evaluated on the vanilla test dataset to maintain consistency.

%Show example multiresolution augmented chips

% \usepackage[normalem]{ulem}
\begin{table}[htpb]
\captionsetup{font=small}
\centering
\caption{Per-class average precisions for three experiments (Vanilla, Multi-resolution, and Augmented). Green, yellow, and red filled cells indicate highest, second highest, and lowest APs per row, respectively.}
\small
\resizebox{\textwidth}{!}{%
\begin{tabular}{lccc}
\textbf{}                                             & \textbf{Vanilla}               & \textbf{Multires}              & \textbf{Aug}                   \\ \hline
\multicolumn{1}{l|}{Aircraft Hangar}                  & \cellcolor[HTML]{FFCCC9}0.1698 & \cellcolor[HTML]{9AFF99}0.5270 & \cellcolor[HTML]{FFFFC7}0.3247 \\
\multicolumn{1}{l|}{Barge}                            & \cellcolor[HTML]{FFCCC9}0.1829 & \cellcolor[HTML]{9AFF99}0.3738 & \cellcolor[HTML]{FFFFC7}0.2210 \\
\multicolumn{1}{l|}{Building}                         & \cellcolor[HTML]{FFFFC7}0.4718 & \cellcolor[HTML]{9AFF99}0.5534 & \cellcolor[HTML]{FFCCC9}0.4451 \\
\multicolumn{1}{l|}{Bus}                              & \cellcolor[HTML]{FFFFC7}0.2949 & \cellcolor[HTML]{9AFF99}0.3773 & \cellcolor[HTML]{FFCCC9}0.2609 \\
\multicolumn{1}{l|}{Cargo Truck}                      & \cellcolor[HTML]{FFFFC7}0.0493 & \cellcolor[HTML]{9AFF99}0.0972 & \cellcolor[HTML]{FFCCC9}0.0445 \\
\multicolumn{1}{l|}{Cargo/container car}              & \cellcolor[HTML]{FFFFC7}0.3659 & \cellcolor[HTML]{9AFF99}0.4737 & \cellcolor[HTML]{FFCCC9}0.1676 \\
\multicolumn{1}{l|}{Cement mixer}                     & \cellcolor[HTML]{FFCCC9}0.0863 & \cellcolor[HTML]{9AFF99}0.1441 & \cellcolor[HTML]{FFFFC7}0.1220 \\
\multicolumn{1}{l|}{Construction site}                & \cellcolor[HTML]{FFFFC7}0.0172 & \cellcolor[HTML]{9AFF99}0.1711 & \cellcolor[HTML]{FFCCC9}0.0032 \\
\multicolumn{1}{l|}{Container crane}                  & \cellcolor[HTML]{FFCCC9}0.0663 & \cellcolor[HTML]{9AFF99}0.2879 & \cellcolor[HTML]{FFFFC7}0.1648 \\
\multicolumn{1}{l|}{Container ship}                   & \cellcolor[HTML]{FFCCC9}0.2269 & \cellcolor[HTML]{9AFF99}0.4660 & \cellcolor[HTML]{FFFFC7}0.3400 \\
\multicolumn{1}{l|}{Crane Truck}                      & \cellcolor[HTML]{9AFF99}0.0946 & \cellcolor[HTML]{FFCCC9}0.0838 & \cellcolor[HTML]{FFFFC7}0.0894 \\
\multicolumn{1}{l|}{Damaged/demolished building}      & \cellcolor[HTML]{FFCCC9}0.0269 & \cellcolor[HTML]{9AFF99}0.0785 & \cellcolor[HTML]{FFFFC7}0.0366 \\
\multicolumn{1}{l|}{Dump truck}                       & \cellcolor[HTML]{FFFFC7}0.1468 & \cellcolor[HTML]{9AFF99}0.2275 & \cellcolor[HTML]{FFCCC9}0.0858 \\
\multicolumn{1}{l|}{Engineering vehicle}              & \cellcolor[HTML]{FFCCC9}0.0020 & \cellcolor[HTML]{9AFF99}0.1234 & \cellcolor[HTML]{FFFFC7}0.0357 \\
\multicolumn{1}{l|}{Excavator}                        & \cellcolor[HTML]{FFFFC7}0.3535 & \cellcolor[HTML]{9AFF99}0.4691 & \cellcolor[HTML]{FFCCC9}0.2064 \\
\multicolumn{1}{l|}{Facility}                         & \cellcolor[HTML]{FFCCC9}0.0777 & \cellcolor[HTML]{9AFF99}0.3750 & \cellcolor[HTML]{FFFFC7}0.1201 \\
\multicolumn{1}{l|}{Ferry}                            & \cellcolor[HTML]{FFCCC9}0.0532 & \cellcolor[HTML]{9AFF99}0.3771 & \cellcolor[HTML]{FFFFC7}0.2197 \\
\multicolumn{1}{l|}{Fishing vessel}                   & \cellcolor[HTML]{FFFFC7}0.1768 & \cellcolor[HTML]{9AFF99}0.1839 & \cellcolor[HTML]{FFCCC9}0.0968 \\
\multicolumn{1}{l|}{Fixed-wing aircraft}              & \cellcolor[HTML]{FFCCC9}0.0888 & \cellcolor[HTML]{9AFF99}0.1218 & \cellcolor[HTML]{FFFFC7}0.1042 \\
\multicolumn{1}{l|}{Flat car}                         & \cellcolor[HTML]{FFCCC9}0.0000 & \cellcolor[HTML]{FFCCC9}0.0000 & \cellcolor[HTML]{FFCCC9}0.0000 \\
\multicolumn{1}{l|}{Front loader/Bulldozer}           & \cellcolor[HTML]{FFCCC9}0.1644 & \cellcolor[HTML]{9AFF99}0.3220 & \cellcolor[HTML]{FFFFC7}0.1959 \\
\multicolumn{1}{l|}{Ground grader}                    & \cellcolor[HTML]{FFFFC7}0.1590 & \cellcolor[HTML]{9AFF99}0.1910 & \cellcolor[HTML]{FFCCC9}0.0289 \\
\multicolumn{1}{l|}{Haul truck}                       & \cellcolor[HTML]{9AFF99}0.3542 & \cellcolor[HTML]{FFCCC9}0.2109 & \cellcolor[HTML]{FFFFC7}0.6875 \\
\multicolumn{1}{l|}{Helicopter}                       & \cellcolor[HTML]{FFFFC7}0.3788 & \cellcolor[HTML]{9AFF99}0.5800 & \cellcolor[HTML]{FFCCC9}0.2965 \\
\multicolumn{1}{l|}{Helipad}                          & \cellcolor[HTML]{FFFFC7}0.2459 & \cellcolor[HTML]{9AFF99}0.4500 & \cellcolor[HTML]{FFCCC9}0.1889 \\
\multicolumn{1}{l|}{Hut/Tent}                         & \cellcolor[HTML]{FFFFC7}0.0004 & \cellcolor[HTML]{9AFF99}0.0006 & \cellcolor[HTML]{FFCCC9}0.0000 \\
\multicolumn{1}{l|}{Locomotive}                       & \cellcolor[HTML]{FFCCC9}0.0760 & \cellcolor[HTML]{9AFF99}0.1929 & \cellcolor[HTML]{FFFFC7}0.1124 \\
\multicolumn{1}{l|}{Maritime vessel}                  & \cellcolor[HTML]{FFCCC9}0.1947 & \cellcolor[HTML]{9AFF99}0.4040 & \cellcolor[HTML]{FFFFC7}0.2884 \\
\multicolumn{1}{l|}{Mobile crane}                     & \cellcolor[HTML]{FFCCC9}0.0248 & \cellcolor[HTML]{9AFF99}0.1375 & \cellcolor[HTML]{FFFFC7}0.0945 \\
\multicolumn{1}{l|}{Motorboat}                        & \cellcolor[HTML]{FFCCC9}0.0811 & \cellcolor[HTML]{9AFF99}0.2488 & \cellcolor[HTML]{FFFFC7}0.1110 \\
\multicolumn{1}{l|}{Oil Tanker}                       & \cellcolor[HTML]{FFCCC9}0.0958 & \cellcolor[HTML]{9AFF99}0.3677 & \cellcolor[HTML]{FFFFC7}0.1193 \\
\multicolumn{1}{l|}{Passenger Vehicle}                & \cellcolor[HTML]{FFFFC7}0.4765 & \cellcolor[HTML]{9AFF99}0.5569 & \cellcolor[HTML]{FFCCC9}0.2980 \\
\multicolumn{1}{l|}{Passenger car}                    & \cellcolor[HTML]{FFFFC7}0.0305 & \cellcolor[HTML]{9AFF99}0.0471 & \cellcolor[HTML]{FFCCC9}0.0000 \\
\multicolumn{1}{l|}{Passenger/cargo plane}            & \cellcolor[HTML]{FFFFC7}0.6508 & \cellcolor[HTML]{9AFF99}0.6691 & \cellcolor[HTML]{FFCCC9}0.6104 \\
\multicolumn{1}{l|}{Pickup Truck}                     & \cellcolor[HTML]{FFFFC7}0.0011 & \cellcolor[HTML]{9AFF99}0.0078 & \cellcolor[HTML]{FFCCC9}0.0000 \\
\multicolumn{1}{l|}{Pylon}                            & \cellcolor[HTML]{FFFFC7}0.0089 & \cellcolor[HTML]{FFCCC9}0.0011 & \cellcolor[HTML]{9AFF99}0.0625 \\
\multicolumn{1}{l|}{Railway vehicle}                  & \cellcolor[HTML]{FFCCC9}0.0000 & \cellcolor[HTML]{9AFF99}0.0833 & \cellcolor[HTML]{FFCCC9}0.0000 \\
\multicolumn{1}{l|}{Reach stacker}                    & \cellcolor[HTML]{FFCCC9}0.0000 & \cellcolor[HTML]{9AFF99}0.2625 & \cellcolor[HTML]{FFCCC9}0.0000 \\
\multicolumn{1}{l|}{Sailboat}                         & \cellcolor[HTML]{9AFF99}0.2614 & \cellcolor[HTML]{FFCCC9}0.0450 & \cellcolor[HTML]{FFFFC7}0.0453 \\
\multicolumn{1}{l|}{Shed}                             & \cellcolor[HTML]{FFCCC9}0.0071 & \cellcolor[HTML]{9AFF99}0.3027 & \cellcolor[HTML]{FFFFC7}0.0277 \\
\multicolumn{1}{l|}{Shipping container}               & \cellcolor[HTML]{FFCCC9}0.0283 & \cellcolor[HTML]{9AFF99}0.3835 & \cellcolor[HTML]{FFFFC7}0.0426 \\
\multicolumn{1}{l|}{Shipping container lot}           & \cellcolor[HTML]{FFCCC9}0.1644 & \cellcolor[HTML]{9AFF99}0.5676 & \cellcolor[HTML]{FFFFC7}0.1890 \\
\multicolumn{1}{l|}{Small aircraft}                   & \cellcolor[HTML]{FFFFC7}0.4610 & \cellcolor[HTML]{FFCCC9}0.3771 & \cellcolor[HTML]{9AFF99}0.4815 \\
\multicolumn{1}{l|}{Small car}                        & \cellcolor[HTML]{FFCCC9}0.3607 & \cellcolor[HTML]{9AFF99}0.4083 & \cellcolor[HTML]{FFFFC7}0.3651 \\
\multicolumn{1}{l|}{Storage Tank}                     & \cellcolor[HTML]{FFCCC9}0.3484 & \cellcolor[HTML]{9AFF99}0.4462 & \cellcolor[HTML]{FFFFC7}0.3700 \\
\multicolumn{1}{l|}{Straddle carrier}                 & \cellcolor[HTML]{FFCCC9}0.3045 & \cellcolor[HTML]{9AFF99}0.4293 & \cellcolor[HTML]{FFFFC7}0.3262 \\
\multicolumn{1}{l|}{Tank car}                         & \cellcolor[HTML]{9AFF99}0.3733 & \cellcolor[HTML]{FFCCC9}0.1123 & \cellcolor[HTML]{FFFFC7}0.2664 \\
\multicolumn{1}{l|}{Tower}                            & \cellcolor[HTML]{FFFFC7}0.0042 & \cellcolor[HTML]{9AFF99}0.1233 & \cellcolor[HTML]{FFCCC9}0.0000 \\
\multicolumn{1}{l|}{Tower crane}                      & \cellcolor[HTML]{FFFFC7}0.0196 & \cellcolor[HTML]{9AFF99}0.0385 & \cellcolor[HTML]{FFCCC9}0.0303 \\
\multicolumn{1}{l|}{Tractor}                          & \cellcolor[HTML]{FFCCC9}0.0000 & \cellcolor[HTML]{9AFF99}0.1109 & \cellcolor[HTML]{FFCCC9}0.0000 \\
\multicolumn{1}{l|}{Trailer}                          & \cellcolor[HTML]{FFCCC9}0.0651 & \cellcolor[HTML]{9AFF99}0.2151 & \cellcolor[HTML]{FFFFC7}0.0861 \\
\multicolumn{1}{l|}{Truck}                            & \cellcolor[HTML]{9AFF99}0.1526 & \cellcolor[HTML]{FFCCC9}0.0469 & \cellcolor[HTML]{FFFFC7}0.1356 \\
\multicolumn{1}{l|}{Truck Tractor}                    & \cellcolor[HTML]{FFFFC7}0.0048 & \cellcolor[HTML]{9AFF99}0.2129 & \cellcolor[HTML]{FFCCC9}0.0000 \\
\multicolumn{1}{l|}{Truck Tractor w/ Box Trailer}     & \cellcolor[HTML]{9AFF99}0.1355 & \cellcolor[HTML]{FFCCC9}0.0863 & \cellcolor[HTML]{FFFFC7}0.1188 \\
\multicolumn{1}{l|}{Truck Tractor w/ Flatbed Trailer} & \cellcolor[HTML]{FFCCC9}0.0322 & \cellcolor[HTML]{9AFF99}0.1261 & \cellcolor[HTML]{FFFFC7}0.0336 \\
\multicolumn{1}{l|}{Truck Tractor w/ Liquid Tank}     & \cellcolor[HTML]{FFFFC7}0.0180 & \cellcolor[HTML]{9AFF99}0.4744 & \cellcolor[HTML]{FFCCC9}0.0000 \\
\multicolumn{1}{l|}{Tugboat}                          & \cellcolor[HTML]{FFFFC7}0.2044 & \cellcolor[HTML]{FFCCC9}0.0380 & \cellcolor[HTML]{9AFF99}0.4119 \\
\multicolumn{1}{l|}{Utility Truck}                    & \cellcolor[HTML]{FFFFC7}0.0098 & \cellcolor[HTML]{9AFF99}0.2846 & \cellcolor[HTML]{FFCCC9}0.0091 \\
\multicolumn{1}{l|}{Vehicle Lot}                      & \cellcolor[HTML]{FFCCC9}0.1105 & \cellcolor[HTML]{9AFF99}0.4115 & \cellcolor[HTML]{FFFFC7}0.1402 \\
\multicolumn{1}{l|}{Yacht}                            & \cellcolor[HTML]{FFFFC7}0.0701 & \cellcolor[HTML]{FFFFC7}0.0701 & \cellcolor[HTML]{9AFF99}0.1868 \\ \hline
\multicolumn{1}{l|}{\textbf{Total mAP}}               & \cellcolor[HTML]{FFCCC9}0.1456 & \cellcolor[HTML]{9AFF99}0.2590 & \cellcolor[HTML]{FFFFC7}0.1549
\end{tabular}
}
\end{table}

%Show cars across geography

%\ffigbox{
%\begin{subfigure}{.23\textwidth}
%  \centering
%  \includegraphics[width=\linewidth]{cars}
%\end{subfigure}
%\begin{subfigure}{.23\textwidth}
%  \centering
%  \includegraphics[width=\linewidth]{planes}
%\end{subfigure}
%}{
%\caption{Left: Extracted cars from four separate geographic locations.  Clear visual differences can be seen.  Right: Extracted planes four geographic locations. They look visually similar and can be clearly distinguished from their background. Imagery in this figure is \copyright 2018 DigitalGlobe, NextView License.}
%}

%Analysis begins here
Training on the multi-resolution dataset produced a better model than both the vanilla and augmented datasets, by a significant margin (0.2590 to 0.1456 and 0.1549 total mean average precision, respectively).  The best detected classes for the multi-resolution dataset are: `passenger/cargo plane', `helicopter', `shipping container lot', `passenger car', and `building'.  The best detected classes for the augmented dataset are: `haul truck', `passenger/cargo plane', `small aircraft', `building', and `tugboat'.  The best detected classes for the vanilla dataset are: `passenger/cargo plane', `passenger car', `building', `small aircraft', and `helicopter'.  The best detected classes bias towards those that have large pixel areas and are set in uniform backgrounds.  For example, most planes are imaged on runways with only slightly changing backgrounds (see Figure 6).  Small cars, even with the second largest instance count, were more poorly detected than rarer but larger and more contextually easy classes. Figure 8 shows extracts of the `small car' class across various geographical contexts.  Several classes were poorly detected across all experiments regardless of average pixel area and instance count.  The `pickup truck' and `hut/tent' classes have over one thousand instances each and yet scored $<$1\% mAP across the board.

There were intra-experimental differences as well as intra-class differences.  The `construction site' class has the largest pixel area at over 373,000 pixels$^2$.  However, the vanilla dataset experiment performed worse than the multi-resolution experiment (0.0172 to 0.1711 AP, respectively).  This significant change could be due to construction sites being cropped too drastically for vanilla (300$^2$) chips, while being more adequately detected by training on the multiple resolution dataset. The `bus' and `tractor' classes each have an area of $<$ 500 pixels$^2$, and were scored significantly higher by the multi-resolution experiment. The overall high-performance of the multi-resolution experiment indicates that the multiple object scales plays an important role in detection precision.  We capped training at seven days for all three experiments to maintain consistency. The poor quality of the augmentation experiment indicates that the augmentations and increased dataset size add significant regularization.  With more training iterations we would expect the augmentation experiment to improve substantially.  Both the multi-resolution and augmented datasets showed continuous performance improvement while the vanilla dataset performance plateaued. 

\vspace{12pt}
\ffigbox{
 \begin{subfigure}{.23\textwidth}
  \centering
  \includegraphics[width=\linewidth]{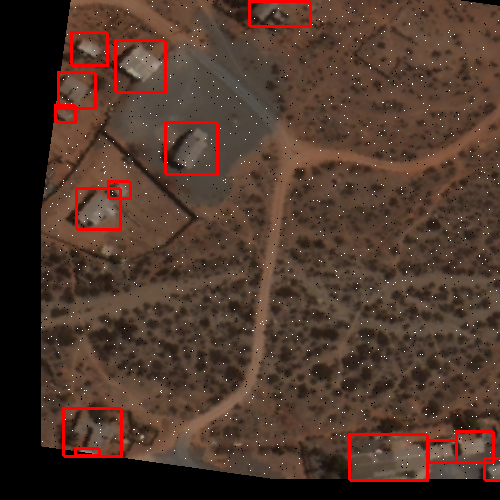}
\end{subfigure}
\begin{subfigure}{.23\textwidth}
  \centering
  \includegraphics[width=\linewidth]{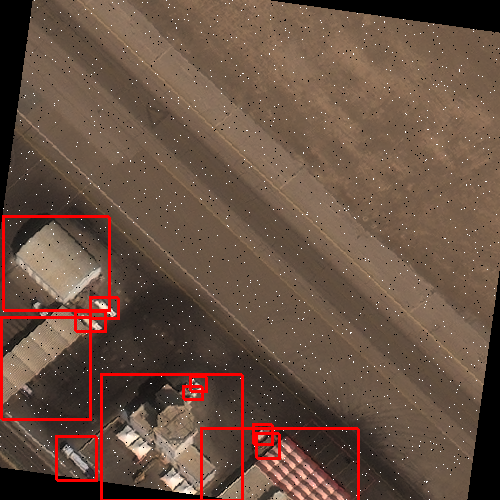}
\end{subfigure}
}{
\caption{Two examples of multi-resolution-augmented chips.  Ground truth bounding boxes are shown in red.  To prevent severe distortion, chip and bounding box rotations were bounded in both directions. Imagery in this figure is from DigitalGlobe. }
}

Our experiments illustrate that the xView dataset is difficult. The SSD meta-architecture with multi-layer feature map extraction achieves relatively low mean average precision. While low instance count and average class size are not directly correlated with per-class average precision, the top detected classes across all experiments bias towards larger and more contextually discernible classes.  Our multi-resolution experiment performed better than both vanilla and augmented datasets.  However, additional training time or additional techniques such as supervised pre-training could improve performance \cite{imagenet_hinton, malik_voc}.

\section{Conclusion and Future Work}
%Conclusion

We introduced xView, an overhead object detection dataset with over 1 million instances in 60 classes.  xView was labeled through an extensive annotation process and three-stage quality control. Our classes include land-use objects such as buildings as well as vehicles and mini-scenes.  xView contains a greater variety of classes and visual contexts than other overhead datasets.  We hope xView will serve as a general-purpose object detection dataset and as a unifying dataset for overhead object detection research. 

%Few-shot learning
There are a variety of future research directions that xView can support.  Few-shot learning is one particular direction that could have broad impacts on overhead imagery applications, e.g., by empowering disaster relief efforts with computer vision tools that are quickly adaptable. The performance of recent few-shot learning techniques \cite{bpl-lake, matchingnets-vinyals} has typically been evaluated on ``K-shot, N-way'' datasets which assume even distributions of instances over categories \cite{omniglot}.  A more realistic benchmark has been proposed by Harihan et al. \cite{lowshot_harihan}, featuring a large base dataset with many instances per category coupled with a low-instance-count target set. xView can be couched in this realistic setting by splitting the categories into high vs. low instance counts. Future extensions to xView could seek to bridge the divide between ``K-shot, N-way'' classification and realistically imbalanced datasets by adding many more object classes with low instance counts (e.g., 20 to 100 instances per category).

%Domain Adaptation
Domain adaptation is another research direction that xView is well-suited to support given the geographic diversity of xView images. The next hurricane, earthquake, or other natural disaster is likely to occur in an area where curated training data (i.e., pre-labeled imagery) is not readily available.  Better domain adaptation techniques are needed to quickly apply existing models to new geographic areas that were not reflected in the training set.

We hope that the release of xView and related code will support future object detection research in the overhead imagery domain as well as in relation to natural imagery.  The RGB release of xView is a general object detection dataset and as such can be used as a standalone object detection benchmark.  We also hope to see xView applied to the intersection of satellite imagery and computer vision, whether for demographic studies or humanitarian efforts.

\section{Appendix Overview}
In the appendix we provide additional dataset examples, comparisons, and examples of images that were re-labeled.

\subsection{Appendix I: Quality Control}

Figure 9 illustrates two examples of pre- and post-quality control images.  Each example is only of a small subset of the overall image.  Images that failed gold-standard quality control were sent for re-labeling.  The two pre-QC examples shown below contain poorly labeled objects or objects unlabeled altogether.  The post-QC examples were remediated.

\subsection{Appendix II: Dataset Examples}

Figure 10 illustrates a fully annotated image.  Elongated objects such cargo ships are often unavoidably enclosed by bounding boxes that contain significant background.  Even seemingly rectangular objects such as shipping crates can have this problem due to the rotation of the objects relative to the satellites and our axis-aligned bounding boxes.  

Figure 11 shows a comparison of chips between xView, COWC, and SpaceNet. Cars are most visible in COWC images because of their aerial perspective.  SpaceNet has the least visible cars. Buildings are the most prominent visual feature in SpaceNet chips. The low sun-angle in the xView chip can be identified through the long object shadows. Table 2 categorizes all 60 classes by their parent-child relationship.  There are 7 parent classes and several classes that do not belong to a parent.  Not all parent classes have equal numbers of fine grained classes.  Figure 12 displays the geographical locations of xView, COWC, and SpaceNet.  A dot was plotted for every region data was captured from, and dots are not scaled based on instance count.  Figure 13 displays an extraction from each class in xView. 

\clearpage

\begin{figure*}[b]
\includegraphics[width=\textwidth]{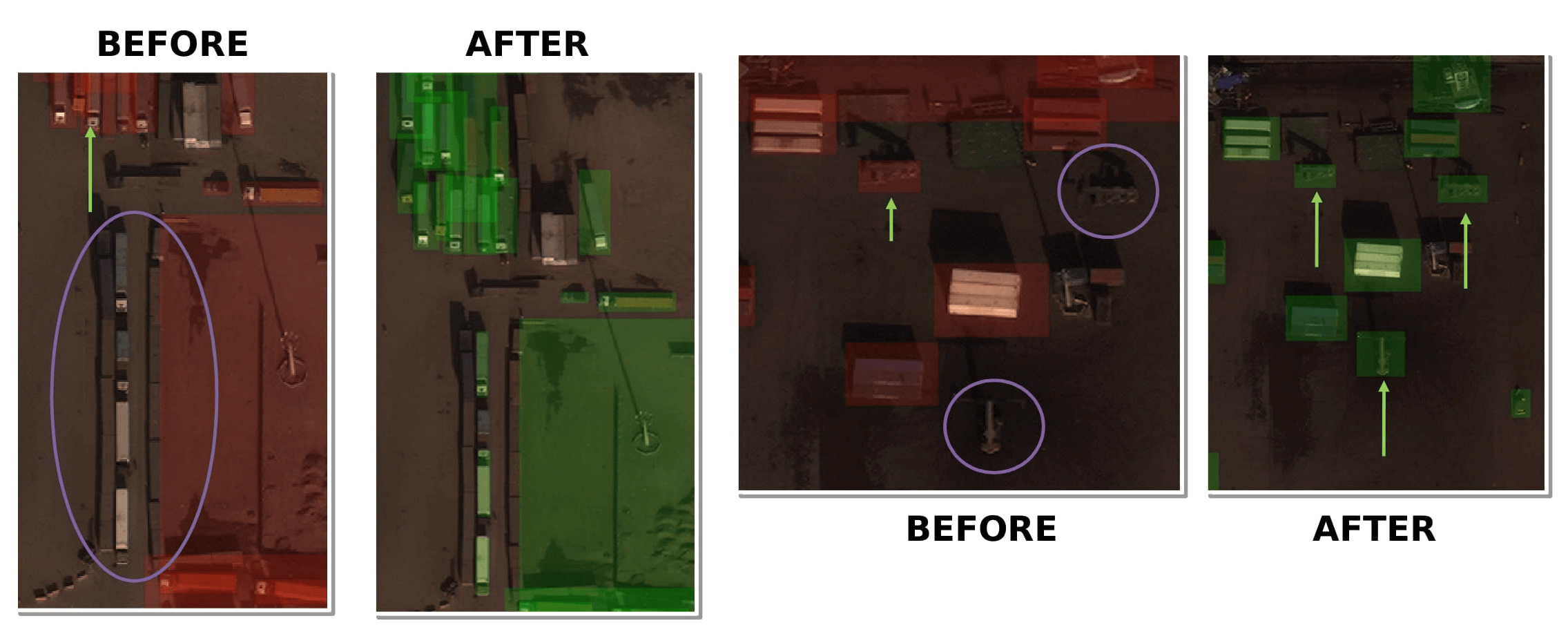}
\caption{Two examples of image chips pre- and post- quality control. Objects that were not labeled or wrongly labeled were remediated.}
\end{figure*}

\begin{figure*}
\includegraphics[width=\textwidth]{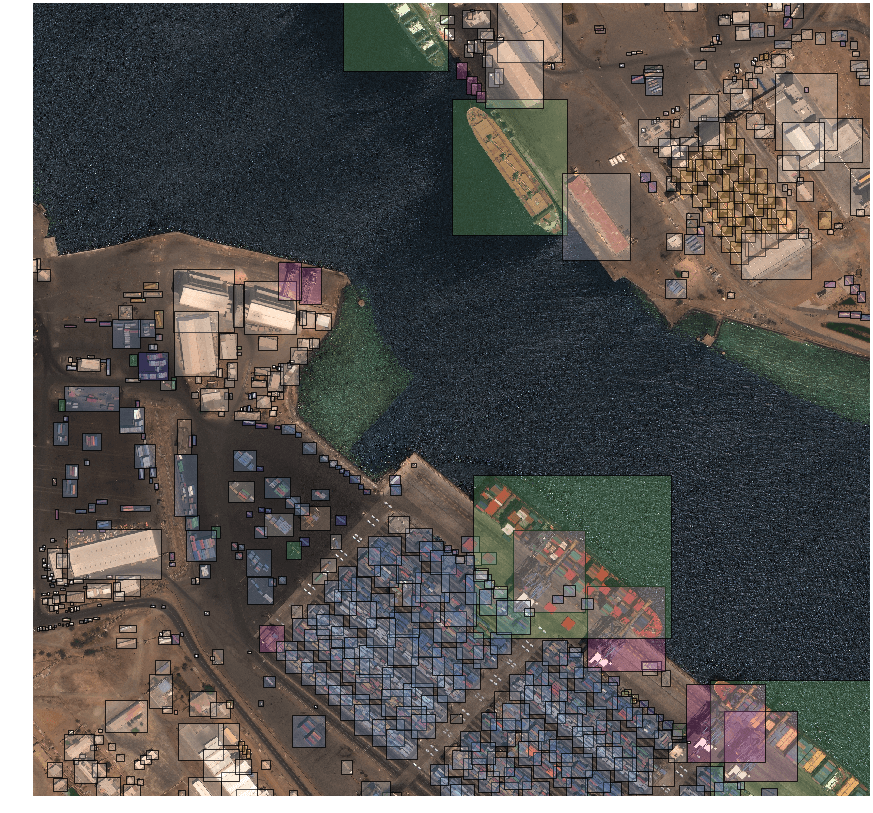}
\caption{A fully annotated image from xView.  Classes are denoted by different bounding box shadings. All imagery in this figure is from Digital Globe.}
\end{figure*}

\begin{figure*}
\centering
\includegraphics[width=\linewidth]{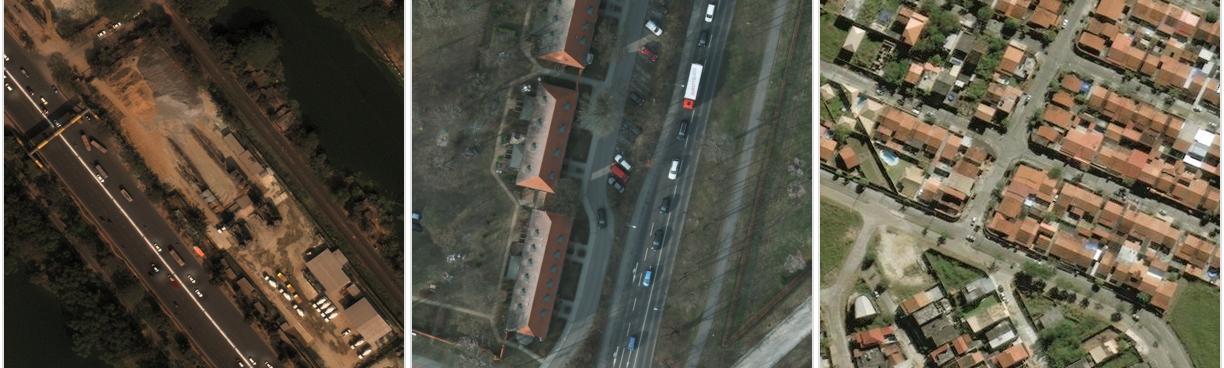}
\caption{xView, COWC, and SpaceNet chips, respectively.  xView imagery in this figure is from DigitalGlobe.}
\end{figure*}

\begin{figure*}
\centering
\includegraphics[width=\linewidth]{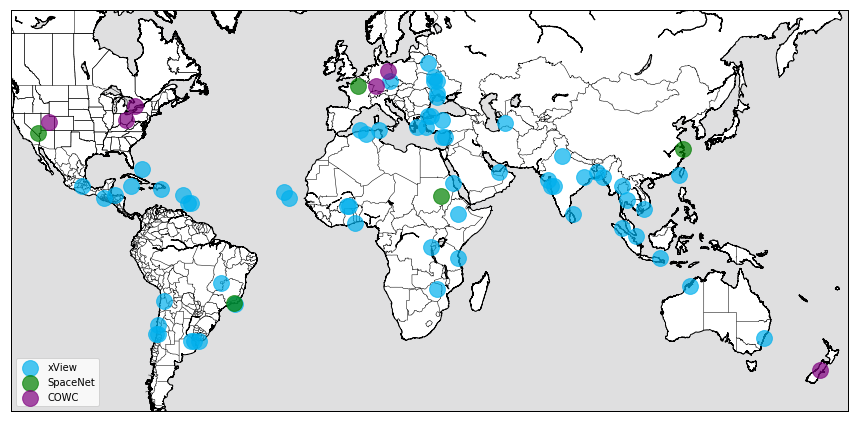}
\caption{The geographical locations of xView, COWC, and SpaceNet. Blue, green, and purple dots represent xView, SpaceNet, and COWC locations respectively.  Dots are not scaled based on instance count.}
\end{figure*}

\begin{table*}[]
\centering
\caption{Parent and child denominations for all 60 classes.  Parent classes are at the headings of each column.  The only exception is the last column `None', which corresponds to classes that have no parent.   }
\large
\resizebox{\textwidth}{!}{
\begin{tabular}{cccccccc}
\textbf{Fixed-Wing Aircraft} & \textbf{Passenger Vehicle} & \textbf{Truck}  & \textbf{Railway Vehicle} & \textbf{Maritime Vessel} & \textbf{Engineering Vehicle} & \textbf{Building} & \textbf{None}          \\ \hline
Small Aircraft               & Small Car                  & Pickup Truck    & Passenger Car            & Motoboat                 & Tower Crane                  & Hut/Tent          & Helipad                \\
Cargo Plane                  & Bus                        & Utility Truck   & Cargo Car                & Sailboat                 & Container Crane              & Shed              & Pylon                  \\
                             &                            & Cargo Truck     & Flat Car                 & Tugboat                  & Reach Stacker                & Aircraft Hangar   & Shipping Container     \\
                             &                            & Truck w/Box     & Tank Car                 & Barge                    & Straddle Carrier             & Damaged Building  & Shipping Container Lot \\
                             &                            & Truck Tractor   & Locomotive               & Fishing Vessel           & Mobile Crane                 & Facility          & Storage Tank           \\
                             &                            & Trailer         &                          & Ferry                    & Dump Truck                   &                   & Vehicle Lot            \\
                             &                            & Truck w/Flatbed &                          & Yacht                    & Haul Truck                   &                   & Construction Site      \\
                             &                            & Truck w/Liquid  &                          & Container Ship           & Scraper/Tractor              &                   & Tower Structure        \\
                             &                            &                 &                          & Oil Tanker               & Front Loader                 &                   & Helicopter             \\
                             &                            &                 &                          &                          & Excavator                    &                   &                        \\
                             &                            &                 &                          &                          & Cement Mixer                 &                   &                        \\
                             &                            &                 &                          &                          & Ground Grader                &                   &                        \\
                             &                            &                 &                          &                          & Crane Truck                  &                   &                       
\end{tabular}
}
\end{table*}

\begin{figure*}
\centering
\includegraphics[width=\linewidth]{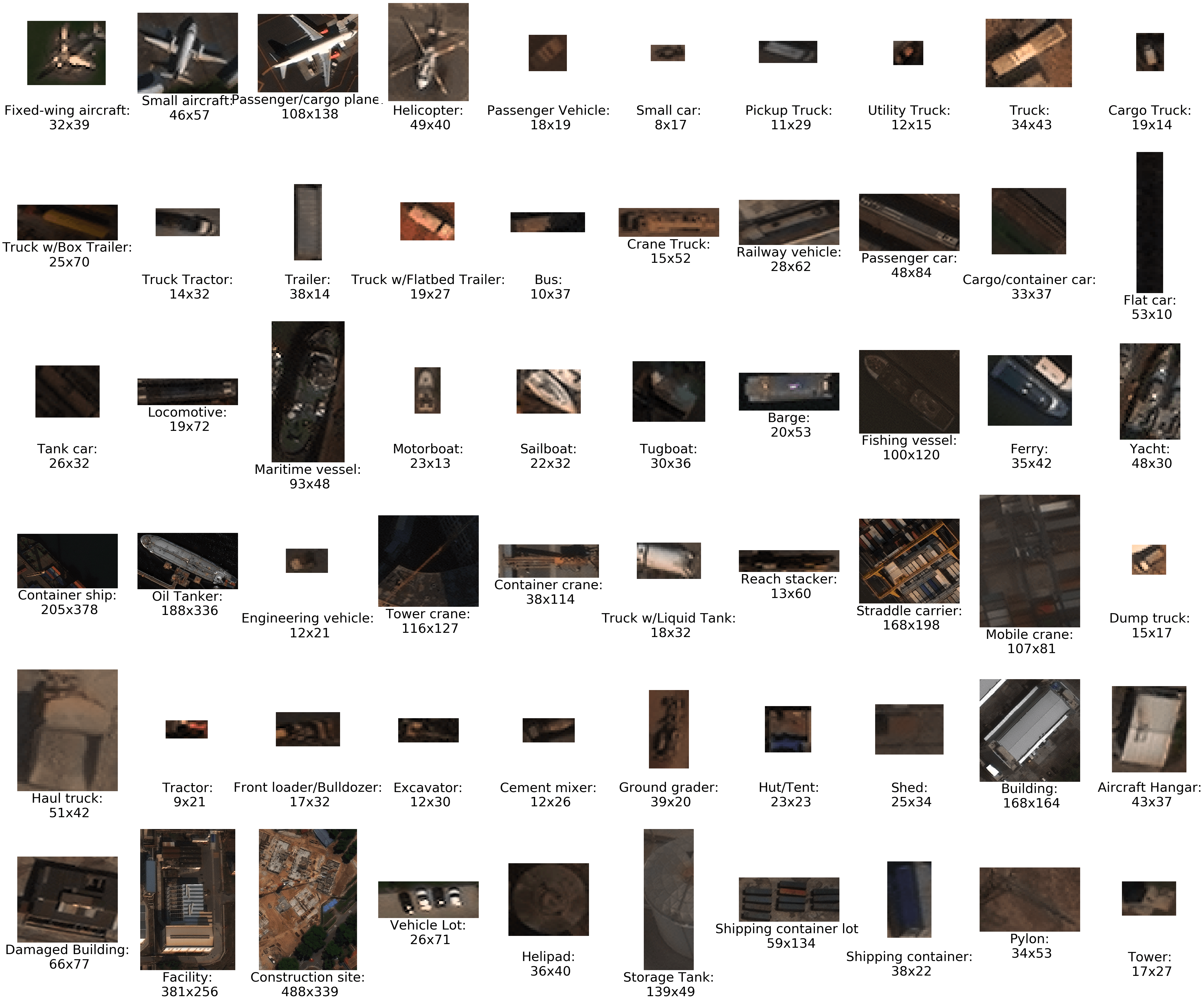}
\caption{An image example per category.  Class labels as well as dimensions for the corresponding image are present under each extraction.  All imagery in this figure is from Digital Globe.}
\end{figure*}

\clearpage
\newpage
\small
\bibliographystyle{ieee}
\bibliography{egbib}

\begin{thebibliography}{10}\itemsep=-1pt

\bibitem{Albert}
A.~Albert, J.~Kaur, and M.~C. Gonzalez.
\newblock Using convolutional networks and satellite imagery to identify
  patterns in urban environments at a large scale.
\newblock In {\em Proceedings of the 23rd ACM SIGKDD International Conference
  on Knowledge Discovery and Data Mining}, KDD '17, pages 1357--1366, New York,
  NY, USA, 2017. ACM.

\bibitem{Azayev}
T.~Azayev.
\newblock Object detection in high resolution satellite images.
\newblock {\em Faculty of Electrical Engineering; Department of Cybernetics},
  2016.

\bibitem{zhaowei}
Z.~Cai, Q.~Fan, R.~S. Feris, and N.~Vasconcelos.
\newblock A unified multi-scale deep convolutional neural network for fast
  object detection.
\newblock In {\em Computer Vision - {ECCV} 2016 - 14th European Conference,
  Amsterdam, The Netherlands, October 11-14, 2016, Proceedings, Part {IV}},
  pages 354--370, 2016.

\bibitem{Chen}
C.~Chen, W.~Gong, Y.~Hu, Y.~Chen, and Y.~Ding.
\newblock Learning oriented region- based convolutional neural networks for
  building detection in satellite remote sensing images.
\newblock {\em The International Archives of the Photogrammetry, Remote Sensing
  and Spatial Information Sciences}, 42(1), 2017.

\bibitem{lrcnn}
M.~Chevalier, N.~Thome, M.~Cord, J.~Fournier, G.~Henaff, and E.~Dusch.
\newblock Lr-cnn for fine-grained classification with varying resolution.
\newblock In {\em 2015 IEEE International Conference on Image Processing
  (ICIP)}, pages 3101--3105, Sept 2015.

\bibitem{fmow}
G.~{Christie}, N.~{Fendley}, J.~{Wilson}, and R.~{Mukherjee}.
\newblock {Functional Map of the World}.
\newblock {\em ArXiv e-prints}, Nov. 2017.

\bibitem{rfcn}
J.~Dai, Y.~Li, K.~He, and J.~Sun.
\newblock R-fcn: Object detection via region-based fully convolutional
  networks.
\newblock In {\em NIPS}, 2016.

\bibitem{pascalvoc}
M.~Everingham, L.~Van~Gool, C.~K.~I. Williams, J.~Winn, and A.~Zisserman.
\newblock The pascal visual object classes (voc) challenge.
\newblock {\em International Journal of Computer Vision}, 88(2):303--338, June
  2010.

\bibitem{Fang2015}
W.~Fang, J.~Chen, C.~Liang, X.~Wang, Y.~Nan, R.~Hu, and S.~Luo.
\newblock Object detection in low-resolution image via sparse representation.
\newblock pages 234--245, 2015.

\bibitem{malik_voc}
R.~Girshick, J.~Donahue, T.~Darrell, and J.~Malik.
\newblock Rich feature hierarchies for accurate object detection and semantic
  segmentation.
\newblock 11 2013.

\bibitem{Hamid}
R.~Hamid, S.~O’Hara, and M.~Tabb.
\newblock Global-scale object detection using satellite imagery.
\newblock {\em The International Archives of the Photogrammetry, Remote Sensing
  and Spatial Information Sciences}, 40(3), 2014.

\bibitem{lowshot_harihan}
B.~Hariharan and R.~B. Girshick.
\newblock Low-shot visual recognition by shrinking and hallucinating features.
\newblock {\em 2017 IEEE International Conference on Computer Vision (ICCV)},
  pages 3037--3046, 2017.

\bibitem{Ishii}
T.~Ishii, E.~Simo-Serra, S.~Iizuka, Y.~Mochizuki, A.~Sugimoto, H.~Ishikawa, and
  R.~Nakamura.
\newblock Detection by classification of buildings in multispectral satellite
  imagery.
\newblock In {\em 2016 23rd International Conference on Pattern Recognition
  (ICPR)}, pages 3344--3349, Dec 2016.

\bibitem{Karlinsky_2017}
L.~Karlinsky, J.~Shtok, Y.~Tzur, and A.~Tzadok.
\newblock Fine-grained recognition of thousands of object categories with
  single-example training.
\newblock In {\em The IEEE Conference on Computer Vision and Pattern
  Recognition (CVPR)}, July 2017.

\bibitem{openimages}
I.~Krasin, T.~Duerig, N.~Alldrin, V.~Ferrari, S.~Abu-El-Haija, A.~Kuznetsova,
  H.~Rom, J.~Uijlings, S.~Popov, A.~Veit, S.~Belongie, V.~Gomes, A.~Gupta,
  C.~Sun, G.~Chechik, D.~Cai, Z.~Feng, D.~Narayanan, and K.~Murphy.
\newblock Openimages: A public dataset for large-scale multi-label and
  multi-class image classification.
\newblock {\em Dataset available from https://github.com/openimages}, 2017.

\bibitem{imagenet_hinton}
A.~Krizhevsky, I.~Sutskever, and G.~E. Hinton.
\newblock Imagenet classification with deep convolutional neural networks.
\newblock {\em Commun. ACM}, 60(6):84--90, May 2017.

\bibitem{bpl-lake}
B.~M. Lake, R.~Salakhutdinov, and J.~B. Tenenbaum.
\newblock Human-level concept learning through probabilistic program induction.
\newblock {\em Science}, 350(6266):1332--1338, 2015.

\bibitem{omniglot}
B.~M. Lake, R.~Salakhutdinov, and J.~B. Tenenbaum.
\newblock Human-level concept learning through probabilistic program induction.
\newblock {\em Science}, 350(6266):1332--1338, 2015.

\bibitem{hongyang}
H.~Li, Y.~Liu, W.~Ouyang, and X.~Wang.
\newblock Zoom out-and-in network with map attention decision for region
  proposal and object detection.
\newblock 09 2017.

\bibitem{coco}
T.-Y. Lin, M.~Maire, S.~Belongie, J.~Hays, P.~Perona, D.~Ramanan,
  P.~Doll{\'a}r, and C.~L. Zitnick.
\newblock Microsoft coco: Common objects in context.
\newblock In D.~Fleet, T.~Pajdla, B.~Schiele, and T.~Tuytelaars, editors, {\em
  Computer Vision -- ECCV 2014}, pages 740--755, Cham, 2014. Springer
  International Publishing.

\bibitem{spacenet}
D.~Lindenbaum.
\newblock Spacenet on aws.

\bibitem{ssd}
W.~Liu, D.~Anguelov, D.~Erhan, C.~Szegedy, S.~Reed, C.-Y. Fu, and A.~C. Berg.
\newblock Ssd: Single shot multibox detector.
\newblock 2016.
\newblock To appear.

\bibitem{cowc}
T.~N. Mundhenk, G.~Konjevod, W.~A. Sakla, and K.~Boakye.
\newblock A large contextual dataset for classification, detection and counting
  of cars with deep learning.
\newblock In B.~Leibe, J.~Matas, N.~Sebe, and M.~Welling, editors, {\em
  Computer Vision -- ECCV 2016}, pages 785--800, Cham, 2016. Springer
  International Publishing.

\bibitem{yolo}
J.~Redmon, S.~Divvala, R.~Girshick, and A.~Farhadi.
\newblock You only look once: Unified, real-time object detection.
\newblock In {\em 2016 IEEE Conference on Computer Vision and Pattern
  Recognition (CVPR)}, pages 779--788, June 2016.

\bibitem{fasterrcnn}
S.~Ren, K.~He, R.~Girshick, and J.~Sun.
\newblock Faster r-cnn: Towards real-time object detection with region proposal
  networks.
\newblock In {\em Proceedings of the 28th International Conference on Neural
  Information Processing Systems - Volume 1}, NIPS'15, pages 91--99, Cambridge,
  MA, USA, 2015. MIT Press.

\bibitem{imagenet}
O.~Russakovsky, J.~Deng, H.~Su, J.~Krause, S.~Satheesh, S.~Ma, Z.~Huang,
  A.~Karpathy, A.~Khosla, M.~Bernstein, A.~C. Berg, and L.~Fei-Fei.
\newblock {ImageNet Large Scale Visual Recognition Challenge}.
\newblock {\em International Journal of Computer Vision (IJCV)},
  115(3):211--252, 2015.

\bibitem{Turner2015}
J.~T. Turner, K.~M. Gupta, B.~T. Morris, and D.~W. Aha.
\newblock Keypoint density-based region proposal for fine-grained object
  detection and classification using regions with convolutional neural network
  features.
\newblock {\em CoRR}, abs/1603.00502, 2015.

\bibitem{matchingnets-vinyals}
O.~Vinyals, C.~Blundell, T.~Lillicrap, K.~Kavukcuoglu, and D.~Wierstra.
\newblock Matching networks for one shot learning.
\newblock In {\em NIPS}, 2016.

\bibitem{zhang07}
W.~Zhang, G.~J. Zelinsky, and D.~Samaras.
\newblock Real-time accurate object detection using multiple resolutions.
\newblock {\em 2007 IEEE 11th International Conference on Computer Vision},
  pages 1--8, 2007.

\bibitem{Zhao2017}
B.~Zhao, J.~Feng, X.~Wu, and S.~Yan.
\newblock A survey on deep learning-based fine-grained object classification
  and semantic segmentation.
\newblock {\em International Journal of Automation and Computing},
  14(2):119--135, Apr 2017.

\end{thebibliography}

\end{document}